\documentclass{article}

\usepackage{PRIMEarxiv}

\usepackage[utf8]{inputenc} 
\usepackage[T1]{fontenc}    
\usepackage{hyperref}       
\usepackage{url}            
\usepackage{booktabs}       
\usepackage{amsfonts}       
\usepackage{nicefrac}       
\usepackage{microtype}      
\usepackage{lipsum}
\usepackage{fancyhdr}       
\usepackage{graphicx}       
\usepackage{float}
\usepackage{footnote}  
\usepackage{natbib}
\usepackage{subcaption}
\usepackage{amsmath}

\graphicspath{{media/}}     

\title{STRMs: Spatial Temporal Reasoning Models for Vision-Based Localization Rivaling GPS Precision
}

\author{
  Hin Wai Lui \\
  Department of Computer Science \\
  University of California \\
  Irvine, CA 92697\\
  \texttt{hwlui@uci.edu} \\
   \And
  Jeffrey L. Krichmar \\
  Department of Computer Science \\
  University of California \\
  Irvine, CA 92697\\
  \texttt{jkrichma@uci.edu} \\
}

\begin{document}
\maketitle


\begin{abstract}
This paper explores vision-based localization through a biologically-inspired approach that mirrors how humans and animals link views or perspectives when navigating their world. We introduce two sequential generative models, VAE-RNN and VAE-Transformer, which transform first-person perspective (FPP) observations into global map perspective (GMP) representations and precise geographical coordinates. Unlike retrieval-based methods, our approach frames localization as a generative task, learning direct mappings between perspectives without relying on dense satellite image databases. We evaluate these models across two real-world environments: a university campus navigated by a Jackal robot and an urban downtown area navigated by a Tesla sedan. The VAE-Transformer achieves impressive precision, with median deviations of 2.29m (1.37\% of environment size) and 4.45m (0.35\% of environment size) respectively, outperforming both VAE-RNN and prior cross-view geo-localization approaches. Our comprehensive Localization Performance Characteristics (LPC) analysis demonstrates superior performance with the VAE-Transformer achieving an AUC of 0.777 compared to 0.295 for VIGOR 200 and 0.225 for TransGeo, establishing a new state-of-the-art in vision-based localization. In some scenarios, our vision-based system rivals commercial smartphone GPS accuracy (AUC of 0.797) while requiring 5$\times$ less GPU memory and delivering 3$\times$ faster inference than existing methods in cross-view geo-localization. These results demonstrate that models inspired by biological spatial navigation can effectively memorize complex, dynamic environments and provide precise localization with minimal computational resources.
\end{abstract}    
\section{Introduction}
\label{sec:intro}

Humans and other animals possess a remarkable ability to navigate familiar environments using first-person perspective (FPP) observations. By simply observing our surroundings, we can determine our location and plan routes to reach desired destinations. This process often requires mentally transforming our viewpoint from FPP to a broader, bird's-eye perspective, what we will refer to as the global map perspective (GMP). Research suggests that humans and many animals naturally convert spatial information between FPP and GMP representations, primarily through vision-based memorization \cite{alexander2015, ulanovsky2007}.

Previous research has modeled this biological capability using VAEs  \cite{xing2022}. These models have been shown to develop internal representations analogous to spatial cell types observed in neuroimaging studies. The latent spaces of these models developed representations resembling place cells \cite{okeefe1971}, border cells \cite{solstad2008, lever2009}, and head direction cells \cite{taube1990} when trained to reconstruct GMP images from FPP sequences and vice versa \cite{xing2022}. This suggests that similar computational principles might underlie both artificial networks and biological spatial cognition.

Open questions remains whether similar principles can be applied to achieve vision-based location memorization in complex, dynamic real-world environments. That is, can a robot equipped with such a model learn to reconstruct GMP images from sequences of FPP images and provide accurate localization in settings with heavy pedestrian and vehicular traffic? Furthermore, how does the precision of such a vision-based approach compare to conventional localization methods like GPS?

To address these questions, we trained two vision-based sequential models, VAE-RNN and VAE-Transformer, across two real-world environments of different scales. The first experiment was conducted at a university campus, where a Jackal robot navigated a 170-meter diagonal area with heavy pedestrian traffic. The second experiment, in a busy urban downtown, involved a Tesla Model 3 traversing a 1.3km diagonal area with both pedestrian and vehicular traffic. High-precision ground truth was obtained using RTK-GPS, which enhances GPS accuracy through atmospheric correction from fixed ground stations.

Unlike previous cross-view localization approaches that rely on retrieval-based methods \cite{tian2017, zhai2017, shi2019spatial, zhu2021vigor, zhu2022transgeo}, our approach frames localization as a generative task: rather than matching an FPP image to an existing satellite image, our model reconstructs an aligned satellite image (GMP) directly from real-time camera feeds (FPP) . This generative approach eliminates the dependency on dense satellite image databases and potentially offers more precise localization by learning the transformation between perspectives. The contributions of this paper are as follows:
\begin{enumerate}
    \item Introduces Spatial Temporal Reasoning Models (STRMs), a biologically-inspired framework that generates place-permanent representations for robust localization.
    
    \item Contributes a new RTK-GPS dataset with centimeter-level spatial precision and high temporal resolution (1Hz) across multiple environment scales.
    
    \item Superior localization measurements with Area Under Curve (AUC) than SoTA given meter deviation thresholds (0.777 vs. 0.295 for VIGOR 200 \cite{zhu2021vigor}) 
    
    \item 5$\times$ less GPU memory and 3$\times$ faster inference than SoTA.
    
    \item Our models rival smartphone GPS accuracy (AUC of 0.777 vs. 0.797 for Phone GPS) while maintaining real-time performance (10.9 FPS) on modest CPU hardware.
\end{enumerate}
\section{Related Work}
\label{sec:related_work}

\subsection{Cross-view Geo-localization}

Cross-view geo-localization determines the geographical location of a ground-level image by matching it with geo-tagged aerial or satellite imagery, confronting challenges of viewpoint, appearance, and scale differences.

Early approaches utilized handcrafted features and geometric alignment \cite{bansal2011geo, lin2013cross}. Deep learning methods later advanced the field, with pioneering work by \citet{workman2015wide} on the CVUSA dataset, followed by improvements from \citet{vo2016localizing} incorporating orientation information. \citet{hu2018cvm} introduced CVM-Net using NetVLAD \cite{arandjelovic2016netvlad} for effective cross-view feature encoding.

Orientation encoding proved critical, with \citet{liu2019lending} using additional input channels and \citet{shi2019spatial} proposing Spatial-aware Feature Aggregation (SAFA) with polar transformation for better alignment. \citet{shi2020optimal} extended this with the Cross-View Feature Transport (CVFT) network.

Several works \cite{regmi2019bridging, toker2021coming, regmi2019cross, tang2019multi} employed generative models for cross-view image synthesis as an auxiliary task, similar to our reconstruction objective, but still relied on retrieval for final localization. A significant limitation of early retrieval-based methods was their dependency on the spatial sampling density of reference images, constraining meter-level accuracy. \citet{zhu2021vigor} addressed this by introducing the VIGOR dataset and a joint retrieval-and-calibration framework, enabling more precise localization by complementing retrieval with regression.

Transformer architectures have shown promise, with \citet{zhu2022transgeo} introducing TransGeo and \citet{yang2021crosslayer} proposing Layer-to-Layer Transformer (L2LTR) leveraging self-attention. For sequence models, \citet{zhang2023cross} created a dataset of FPP images with aerial views, though still using retrieval rather than generative regression.

Recent innovations include contrastive learning approaches like \citet{deuser2023sample4geo}'s Sample4Geo using symmetric InfoNCE loss, as well as part-based representation learning with \citet{wang2021each}'s Local Pattern Networks (LPN) focusing on distinctive image parts.

While most existing methods focus on cross-view matching or retrieval, our generative approach directly transforms first-person perspective observations into global map perspective representations and precise coordinates.

\subsection{Camera Relocalization}

Camera relocalization is the task of estimating the 6-DoF camera pose (position and orientation) within a known scene from monocular images, which is essential for applications in augmented reality, robotics navigation, and autonomous driving~\cite{kendall2015posenet, shotton2013scene}. Traditional approaches include image retrieval-based methods~\cite{ding2019camnet}, direct pose regression~\cite{kendall2015posenet}, and scene coordinate regression~\cite{shotton2013scene}.

Among these, PoseNet~\cite{kendall2015posenet} pioneered direct pose regression by adapting a classification CNN (GoogLeNet) to predict 6-DoF camera pose from a single image. This approach demonstrated that deep networks could learn the complex mapping from image appearance to camera pose without requiring explicit 3D scene models or feature matching. Subsequent extensions have improved upon this foundation by introducing geometric loss functions~\cite{kendall2017geometric} and uncertainty modeling~\cite{kendall2016modelling}.

Our work is most closely related to regression approaches like PoseNet, which pioneered the use of convolutional neural networks for camera relocalization in the Cambridge Landmark Dataset, where train and test trajectories are not exactly aligned—similar to our scenario. However, a critical distinction lies in the spatial scale: most outdoor scenes in existing camera relocalization datasets cover relatively small areas, with even the largest Cambridge scene spanning only approximately 60m × 60m. In contrast, our dataset operates at a substantially larger scale (up to 1.3km diagonal), highlighting the fundamental difference between camera relocalization and geo-localization. While the former primarily aims at accurate camera pose estimation for 3D scene reconstruction in confined spaces, the latter focuses on determining precise latitude and longitude coordinates across much wider geographical areas.

\subsection{Visual SLAM}
Visual SLAM (Simultaneous Localization and Mapping) aims to simultaneously determine a robot's location within an unknown environment while building a map of the surroundings using visual sensors. Traditional approaches include feature-based methods like ORB-SLAM \cite{mur2015orb, mur2017orb2}, which extract visual features for tracking, and direct methods such as LSD-SLAM \cite{engel2014lsd} and DTAM \cite{newcombe2011dtam}. More recent approaches have integrated inertial sensors \cite{campos2021orb3}, RGB-D cameras \cite{endres2012rgbd}, and semantic information \cite{son2023sce} to enhance accuracy in various environments. Despite these advancements, most SLAM systems still struggle with global localization in large-scale environments and dynamic scenes.

Unlike conventional Visual SLAM approaches that rely on visual and sensor based odometry to estimate relative pose movement, which introduces substantial error accumulation and drift over time, our method implements a spatial-temporal reasoning model that provides exact latitude and longitude for each frame in the sequence. While traditional SLAM methods construct local maps with inherent scale ambiguities and require loop closure techniques to correct drift, our approach directly learns to transform first-person perspective (FPP) observations into global map perspective (GMP) representations.
\section{Data Collection}

Our work required highly accurate localization ground truth data for training and evaluation. We employed RTK-GPS technology to establish centimeter-level precision for our datasets, collected in two distinct environments using different robotic platforms: a ground robot (Jackal) and an electric vehicle (Tesla Model 3).

\subsection{RTK-GPS for Ground Truth}

Real-Time Kinematic GPS (RTK-GPS) provides centimeter-level positioning accuracy by correcting for atmospheric interference and other error sources that affect standard GPS signals \cite{teunissen2015review}. The system uses a stationary base station at a known location to calculate corrections for GPS signals, which are transmitted via NTRIP over the internet. Mobile devices (rovers) within approximately 10km apply these corrections to achieve centimeter-level accuracy.

For our data collection, we utilized a UNAVCO \cite{UNAVCO} public base station approximately 6km from the university campus and 12km from the downtown area. We connected to this base station using the NTRIP protocol, applying corrections to our SparkFun GPS-RTK-SMA module in real-time. To verify accuracy, we conducted static tests at known locations, confirming position reporting within 1 meter of surveyed coordinates. For quality control, we excluded data points with horizontal accuracy values reported by the RTK-GPS module exceeding 5 meters.

\begin{figure}[H]
    \centering
    \begin{subfigure}{0.48\textwidth}
        \includegraphics[width=\textwidth]{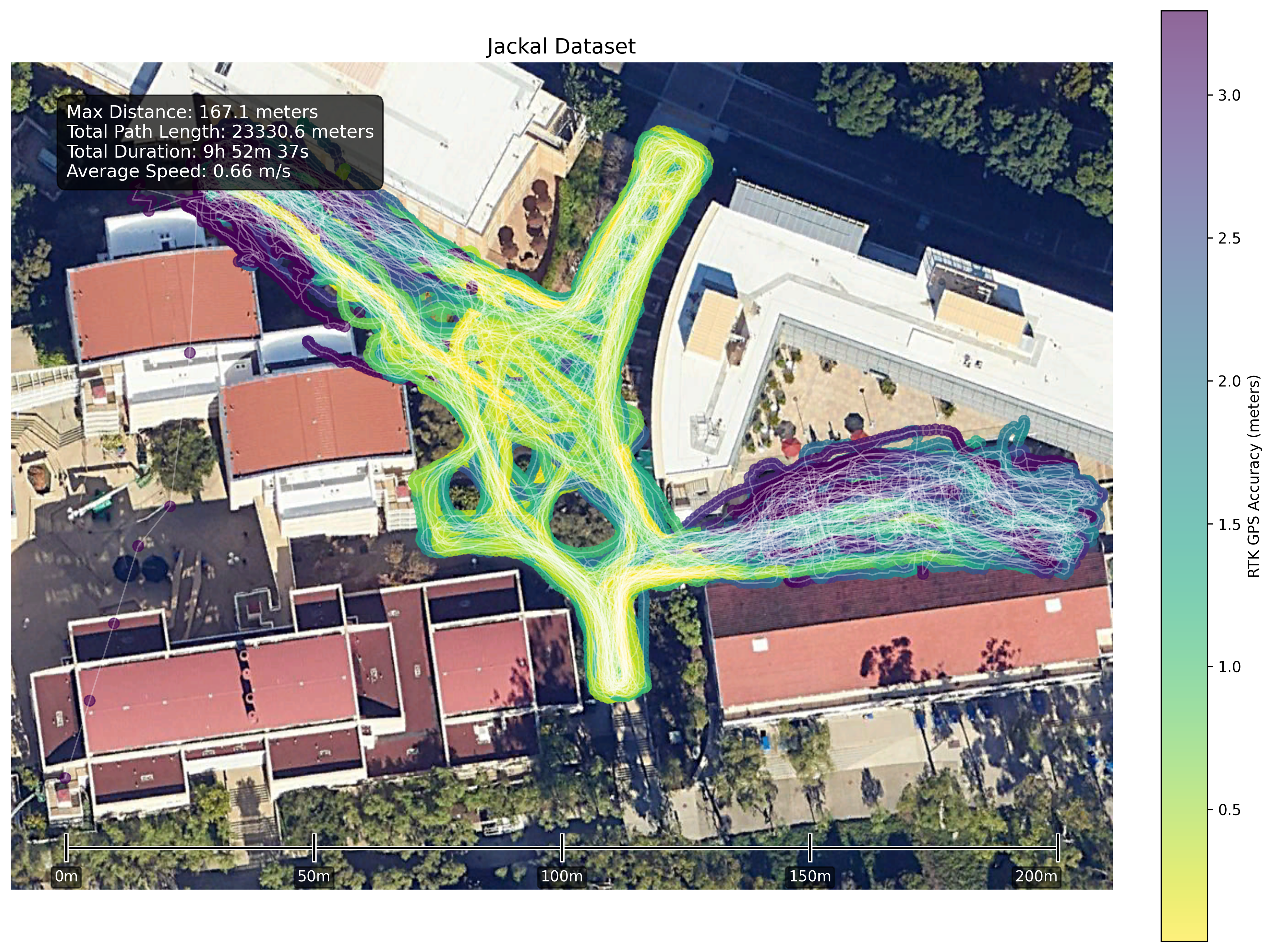}
        \caption{Jackal robot paths. The robot traversed a 170m diagonal area with pedestrian traffic on a university campus.}
        \label{fig:jackal_paths}
    \end{subfigure}
    \hfill
    \begin{subfigure}{0.48\textwidth}
        \includegraphics[width=\textwidth]{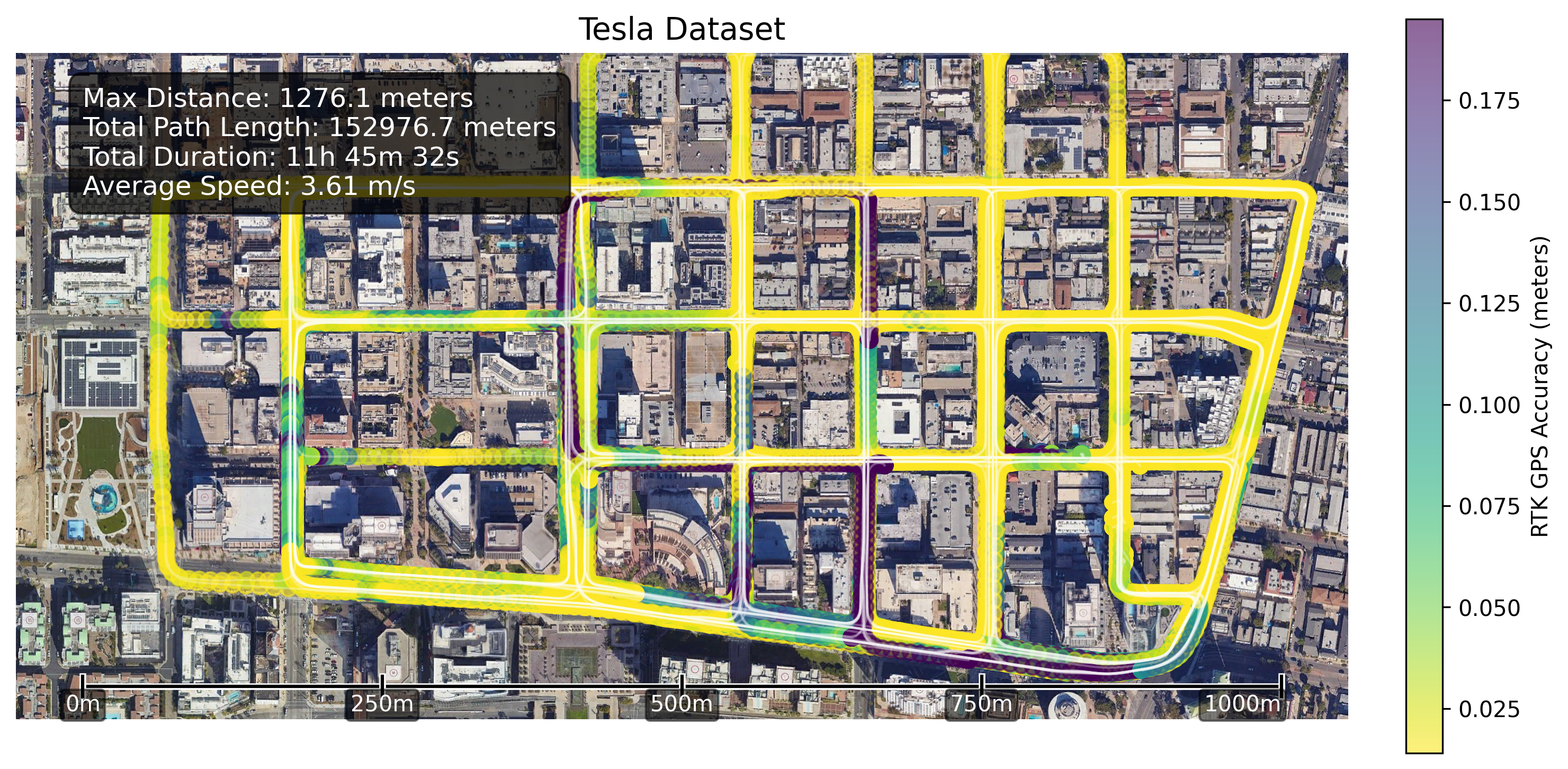}
        \caption{Tesla Model 3 paths. The vehicle navigated a 1.3km diagonal urban area with both pedestrian and vehicular traffic.}
        \label{fig:tesla_paths}
    \end{subfigure}
    \caption{Dataset collection paths for both robotic platforms. Colors indicate RTK-GPS accuracy in meters, with yellow representing higher precision.}
    \label{fig:dataset_paths}
\end{figure}

\subsection{Datasets}
For our smaller-scale environment, we deployed a Clearpath Jackal UGV across a university campus covering approximately 170m diagonally (Figure \ref{fig:jackal_paths}). Data collection occurred over multiple days and times to capture varying conditions. We gathered approximately 10 hours of total video footage synchronized with RTK-GPS location data and IMU readings. Video was recorded with a variable frame rate adjusted according to exposure conditions, with the Jackal robot maintaining an average speed of 0.66 m/s throughout data collection, allowing for detailed capture of environmental features at walking pace.

The Tesla Dataset covered an urban downtown area spanning approximately 1.3km diagonally (Figure \ref{fig:tesla_paths}), featuring dense building structures and typical urban traffic patterns. Approximately 12 hours of data was gathered. The Tesla Model 3 vehicle maintained an average speed of 3.61 m/s (approximately 13 km/h), appropriate for urban driving conditions with traffic signals and pedestrian crossings. Video data was captured at 36 FPS, providing high temporal resolution even at vehicular speeds.

\subsection{Data Processing}
We provide raw video footage and sensor data, along with two data processing scripts that generate the processed datasets. These scripts perform the following steps: (1) extracting frames at regular intervals to balance temporal coverage and computational efficiency, (2) filtering out frames with poor RTK-GPS accuracy, (3) processing camera frames to standardized sizes, (4) generating corresponding satellite images by cropping from high-resolution satellite base images using RTK-GPS coordinates, and (5) storing all images and metadata in an efficient binary format.

Users can apply these scripts to produce datasets containing paired first-person perspective (FPP) images and corresponding global map perspective (GMP) satellite views, along with precise RTK-GPS coordinates and sensor readings.

\section{Methodology}

\subsection{Model Architecture}
\begin{figure}[H]
\centering
\begin{subfigure}{0.5\columnwidth}
\includegraphics[width=\textwidth]{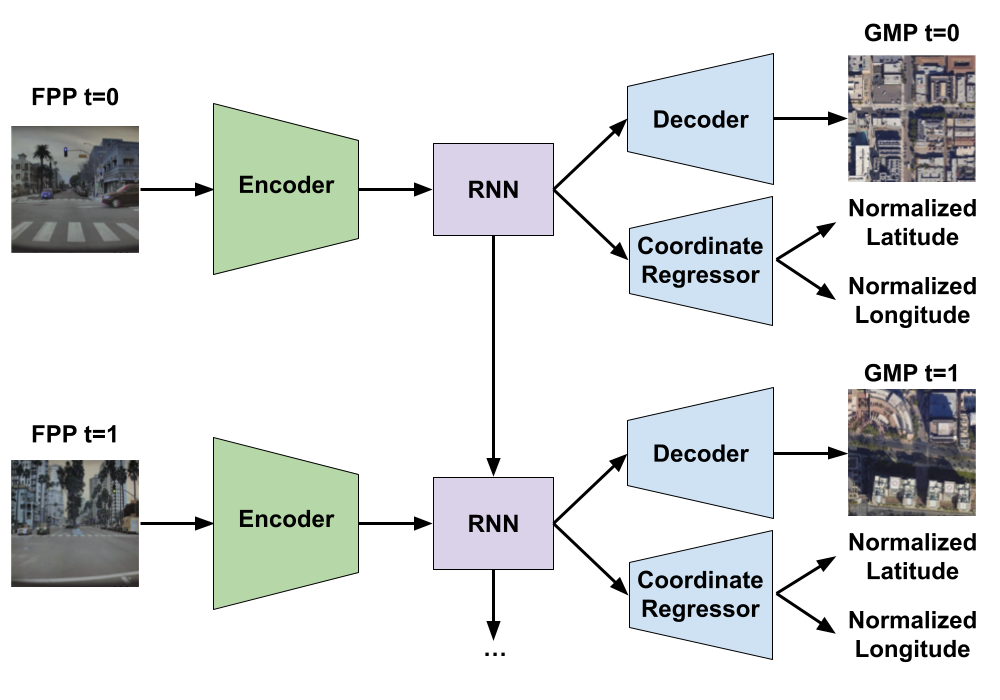}
\caption{VAE-RNN Architecture}
\label{fig:vae_rnn}
\end{subfigure}

\begin{subfigure}{0.5\columnwidth}
\includegraphics[width=\textwidth]{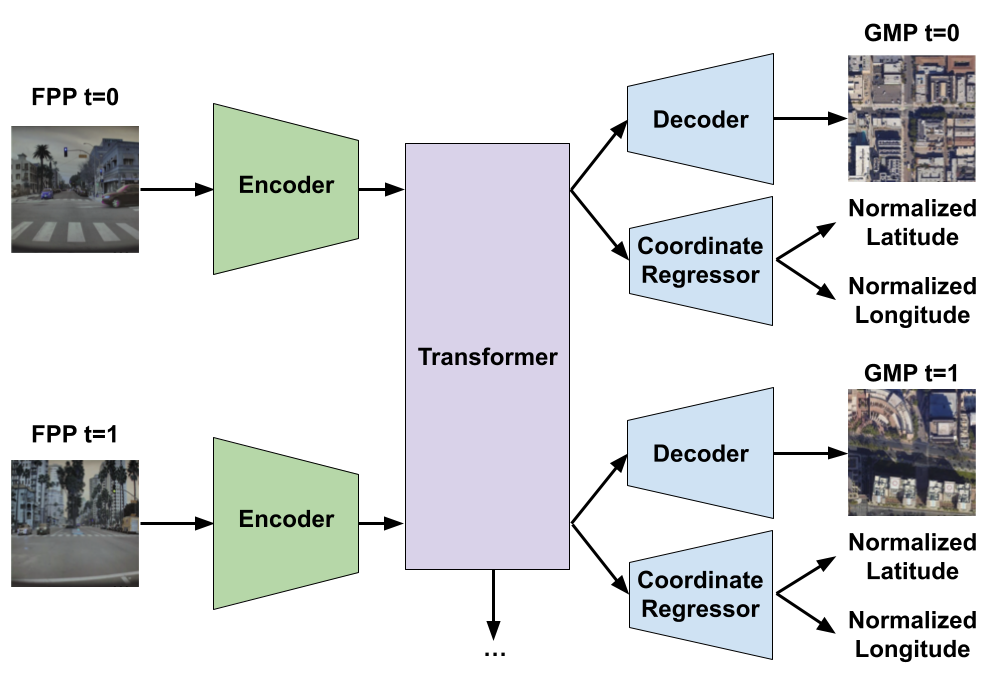}
\caption{VAE-Transformer Architecture}
\label{fig:vae_transformer}
\end{subfigure}
\caption{Model architectures for vision-based localization. Both encode FPP images and generate GMP images and coordinate prediction, but differ in sequential processing: VAE-RNN uses RNNs, while VAE-Transformer uses a causal transformer encoder.}
\label{fig:model_architecture}
\end{figure}

In this work, we develop two spatial-temporal reasoning models for vision based geo-localization: VAE-RNN and VAE-Transformer, illustrated in Figure \ref{fig:model_architecture}.

Both architectures share the same CNN encoder with four convolutional layers (channels [16, 32, 32, 32], filter size 3) and decoder with five transpose convolutional layers (channels [512, 256, 128, 64, 32], filter size 3). The VAE-RNN uses RNN networks to process features sequentially, while the VAE-Transformer employs a causal transformer that processes the entire sequence simultaneously while preserving temporal causality.

A separate MLP regresses coordinates from the latent space to normalized latitude and longitude. The normalization process involves setting predefined fixed boundaries for the maximum and minimum latitude and longitude values.

\subsection{Loss Function}
Our training objective combines three components:

\begin{equation}
\mathcal{L} = \mathcal{L}_{\text{recon}} + \beta(t) \mathcal{L}_{\text{KL}} + \mathcal{L}_{\text{coord}}
\end{equation}

The reconstruction loss ($\mathcal{L}_{\text{recon}}$) is the mean squared error between predicted and ground truth GMP images. The KL divergence loss ($\mathcal{L}_{\text{KL}}$) regularizes the latent space toward a standard normal distribution. The coordinate loss ($\mathcal{L}_{\text{coord}}$) measures the Euclidean distance between predicted and ground truth normalized coordinates. 

To prevent posterior collapse, we implement KL annealing where $\beta(t)$ increases linearly from 0 to 1 during training.

\subsection{Hyperparameter Optimization}
We conducted an extensive hyperparameter search using 10\% of stratified data with 5 separate random seeds for each configuration to determine optimal model configurations. The search explored learning rate, sequence length, time between frames, latent size, and transformer-specific parameters ($d\_model$, $n\_heads$, and $n\_layers$).

\subsection{Training and Inference}
We train using the Adam optimizer with environment-specific learning rates, a batch size of 16, 1,000 latent variables, a 10-second interval between frames, and a sequence length of 24, all determined through hyperparameter search. For the VAE-Transformer, we use 8 layers with 16 attention heads. We employ early stopping with a patience of 5 epochs to prevent overfitting, utilizing a validation set comprising 10\% of the training data.

During inference, the model processes a continuous stream of FPP images by maintaining a buffer of recent frames. Normalized coordinate predictions are converted to real-world measurements using the geographical distances of the bounding box, yielding localization error in meters.

\section{Results}

\subsection{Localization Performance Characteristics (LPC)}

\begin{figure}[H]
\centering
\includegraphics[width=0.48\textwidth]{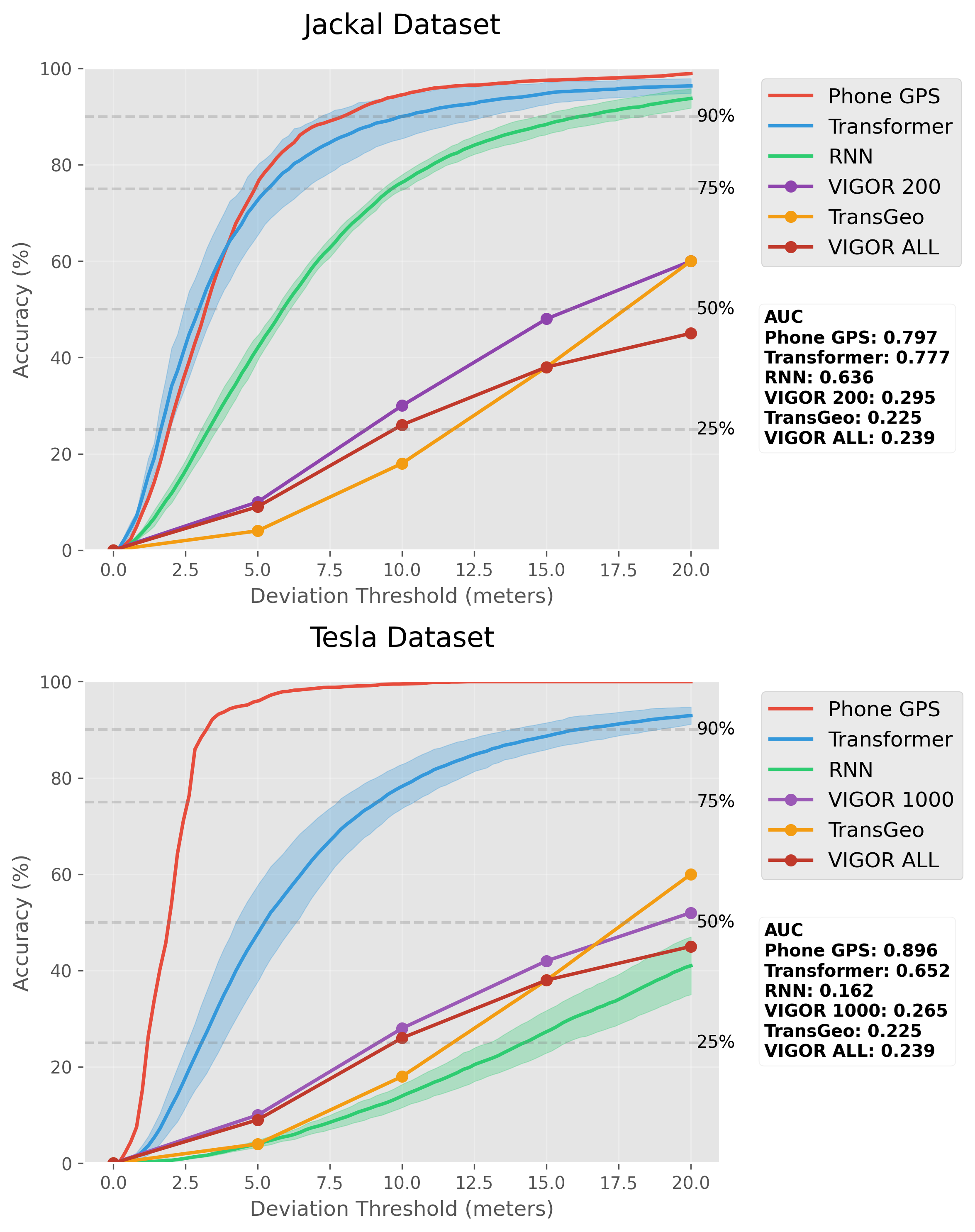}
\caption{Localization performance characteristics (LPC) comparison on Jackal (top) and Tesla (bottom) datasets. Curves show the percentage of accurate localization measurements within a given meter deviation threshold. The color bands of our models represent confidence band across three runs with different random seeds. AUC values are shown for each method, with higher values indicating better overall performance across all thresholds.}
\label{fig:performance_curves}
\end{figure}

To comprehensively evaluate our models, we repeated each experiment three times with different random seeds using the full dataset to establish robust localization performance characteristics (LPC) for both the RNN and Transformer models. In addition, we included the physical GPS measurements from an Android phone (Samsung Galaxy S22 Ultra) for comparison. Figure \ref{fig:performance_curves} presents LPC as Accuracy vs. Deviation Threshold plots for both datasets.

In these plots, accuracy represents the percentage of location measurements that fall within the specified deviation threshold (in meters) from the RTK-GPS ground truth. These plots are analogous to the Receiver Operating Characteristic (ROC) plots in classification literature, with the Area Under the Curve (AUC) serving as a comprehensive metric of each measurement's performance across all thresholds. The colored bands of our model visualize the confidence band of inference results across the three independent runs, illustrating the consistency and robustness of each approach.

Several key observations emerge from this analysis. First, the Transformer consistently outperforms the RNN model across both datasets, demonstrating the superiority of the transformer architecture for this spatial-temporal reasoning task. The performance gap is particularly pronounced in the Tesla dataset.

When comparing our machine learning models against commercial Phone GPS, we observe that the Transformer model in the Jackal dataset achieves LPC comparable to Phone GPS (AUC of 0.777 vs. 0.797). This finding is particularly significant because it demonstrates that a machine learning model can match the performance of a dedicated physical measurement device operating at the exact same point in space and time. The fact that our model maintains this level of accuracy despite being trained on data collected at different times demonstrates that vision-based spatial-temporal reasoning models can generalize effectively across temporal variations in the same environment, such as changing lighting conditions and pedestrian traffic patterns.

We compared our models against state-of-the-art methods in the cross-view geo-localization literature. While prior work often focuses on Top-X accuracy and retrieval hit rates, these metrics do not directly translate to localization precision in meters, as they depend on the density and spatial resolution of satellite image databases. VIGOR \cite{zhu2021vigor} was the first dataset to provide precise latitude and longitude coordinates of the reference satellite images, enabling direct comparison of LPC against our models and physical GPS measurements. We benchmarked against VIGOR's GPS-assisted retrieval scenarios, where the retrieval algorithm restricts the search space to images within a certain radius of an initial GPS measurement. It's worth noting that VIGOR utilizes the same SAFA \cite{shi2019spatial} model architecture as its feature extractor and matching component. We include results from VIGOR-ALL (entire test dataset), VIGOR-200 (200m search radius), and VIGOR-1000 (1000m search radius), manually extracting their LPC from published figures for comparison at different deviation thresholds. We also include TransGeo \cite{zhu2022transgeo}, which uses the same VIGOR dataset but employs a transformer architecture. Our Transformer model significantly outperforms all previous retrieval-based methods, even when those methods artificially restrict their search space to match the predictive limits of our model. These results demonstrate that our generative approach to geo-localization, which directly transforms FPP observations into GMP representations and precise coordinates, offers substantial advantages over traditional retrieval-based methods. By removing the dependency on dense satellite image databases and learning the direct mapping between visual features and geographical coordinates, our model achieves state-of-the-art localization performance in real-world environments.

\subsection{Visualizing Real World Performance}

\begin{figure}[H]
\centering
\includegraphics[width=0.48\textwidth]{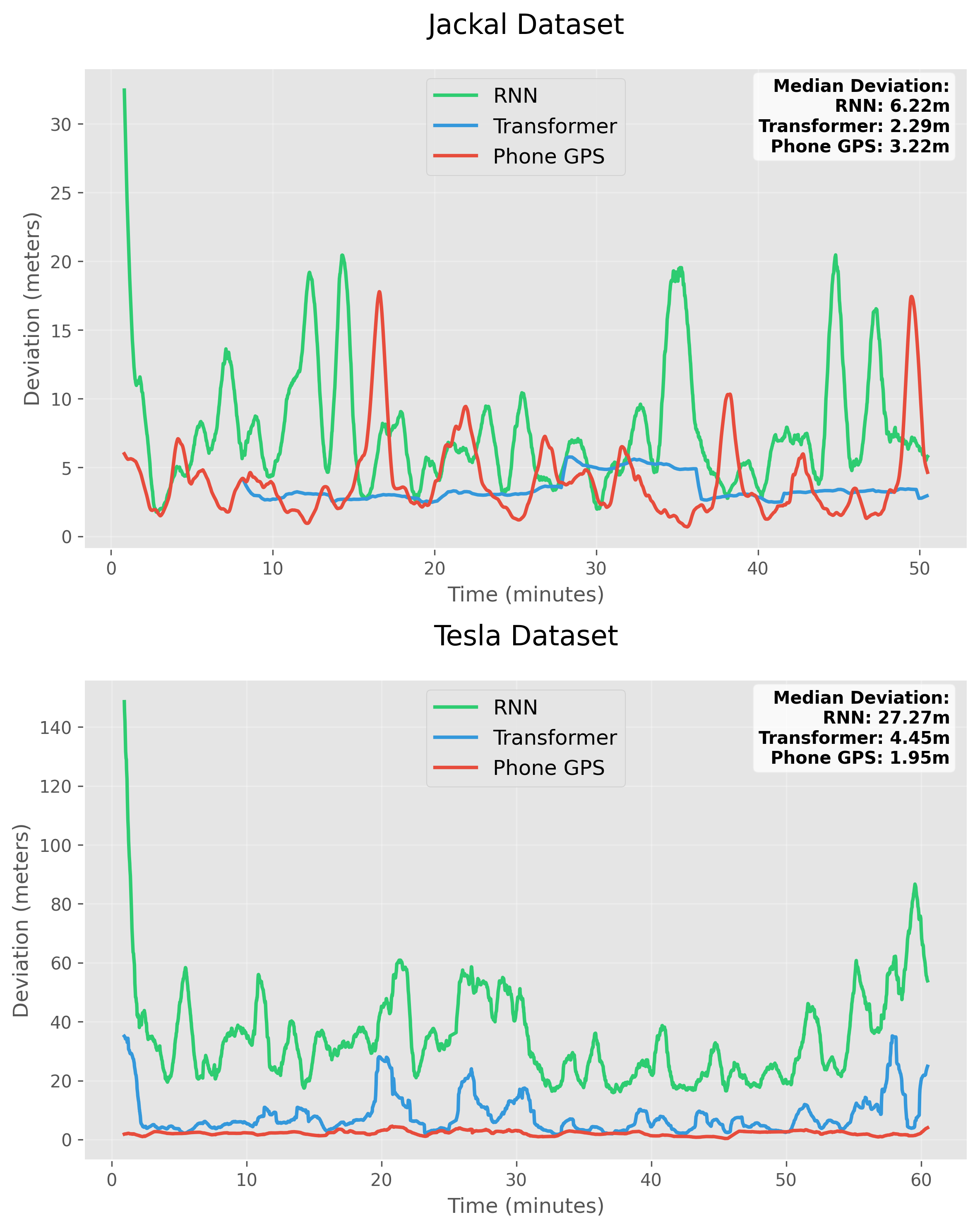}
\caption{Localization performance comparison between VAE-RNN (green), VAE-Transformer (blue), and Phone GPS (red) on Jackal (top) and Tesla (bottom) datasets. The plots show deviation from RTK-GPS ground truth in meters over time.}
\label{fig:localization_comparison}
\end{figure}

Figure \ref{fig:localization_comparison} presents the inference localization performance of VAE-RNN and VAE-Transformer across both test environments in one of the three random seeds. This evaluation represents a realistic real-world test case, as the models processed continuous, uninterrupted video streams. Unlike evaluations that rely on preprocessed, chunked sequences from a dataloader, our inference tests mirrors deployment conditions where systems must handle continuous visual input with its inherent variability, occlusions, lighting changes, etc. Additional visualizations of the test paths are available in the supplementary material.

In the Jackal dataset (university campus), which has corridors between large buildings, the VAE-Transformer achieved a median deviation of 2.29m from ground truth, outperforming both the VAE-RNN (6.22m) and Phone GPS (3.22m). This deviation represents only 1.37\% of the environment's 167m diagonal distance. The Transformer model demonstrates remarkable stability throughout the 50-minute test period, while the RNN exhibits significant fluctuations with spikes exceeding 20m. In the Tesla dataset (urban downtown), the performance gap widens further, with the VAE-Transformer achieving 4.45m median deviation (merely 0.35\% of the environment) compared to the RNN's 27.27m (2.14\%). When we convert the spatial error (4.45m) to temporal terms by dividing by the vehicle's speed (3.61m/s), we find that the median localization error is equivalent to just 1.23 seconds of vehicle movement, which is only slightly larger than our RTK-GPS ground truth sampling interval of 1 second. The RNN model shows extreme instability in this larger environment with multiple deviation spikes above 50m. While Phone GPS performs better in this urban scenario with a 1.95m median deviation, the Transformer's ability to maintain consistent accuracy of less than 5m demonstrates that vision-based localization can approach commercial GPS performance even in complex urban environments.

\subsection{Reconstruction Ablation Analysis}

\begin{figure}[H]
\centering
\includegraphics[width=0.48\textwidth]{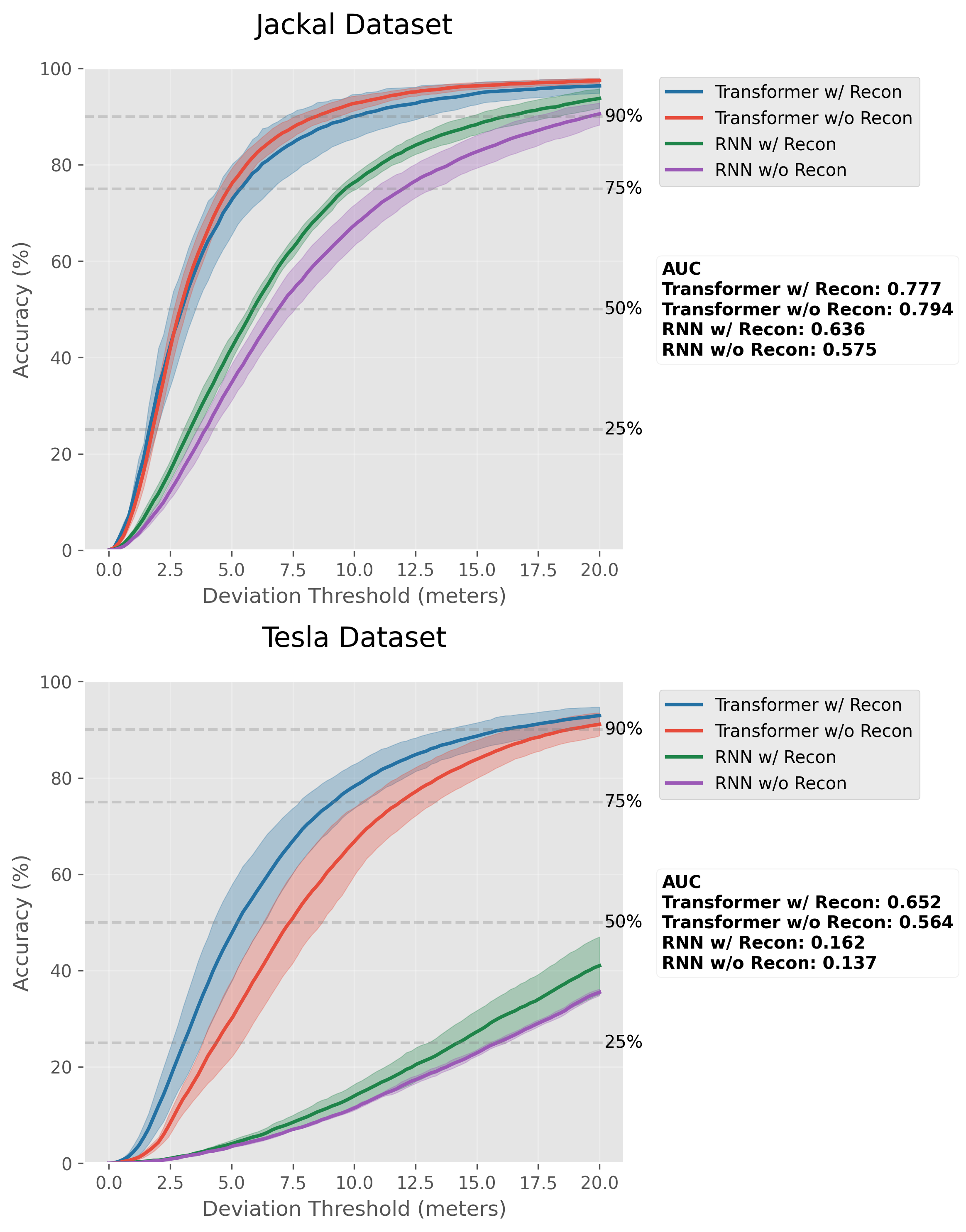}
\caption{Localization performance characteristics comparison between models trained with and without reconstruction (w/ Recon vs. w/o Recon) across both datasets. The ablation shows reconstruction's importance increases with environmental complexity, particularly evident in the Tesla urban downtown dataset where performance drops without it.}
\label{fig:reconstruction_ablation}
\end{figure}

To investigate the effect of perspective transform reconstruction on localization performance, we conducted an ablation study comparing models trained with and without the reconstruction objective. Figure~\ref{fig:reconstruction_ablation} presents the LPC across both datasets and architectures. Removing the reconstruction objective generally degrades performance, with one notable exception: the Transformer model on the Jackal dataset slightly improves without reconstruction (AUC of 0.794 vs. 0.777). However, this performance difference is not statistically significant, as evidenced by the overlapping confidence bands. This exception can be attributed to the simplicity and smaller scale of the Jackal environment, where the Transformer's substantial modelling capacity can effectively model location-specific features without additional regularization from the reconstruction task.

In the more complex Tesla urban downtown environment, perspective transform reconstruction provides crucial regularization that significantly improves localization, with the Transformer's AUC decreasing from 0.652 to 0.564 without it.

\begin{figure}[H]
\centering
\includegraphics[width=0.48\textwidth]{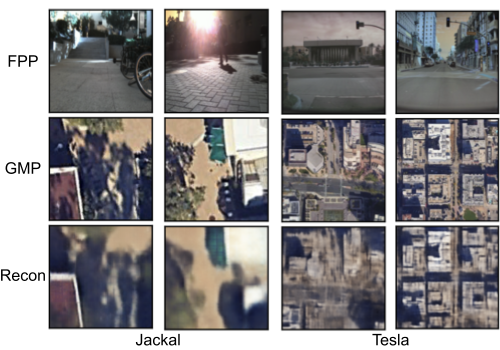}
\caption{Examples of GMP image reconstruction results.}
\label{fig:recon_results}
\end{figure}

Figure \ref{fig:recon_results} shows examples of reconstructed GMP images, demonstrating the model's ability to transform perspectives while preserving location-specific features. This transformation process encourages development of internal representations that capture permanent environmental characteristics while disregarding temporary elements—a capability essential for robust localization in dynamic real-world settings. Additional examples of the reconstructed images can be found in the supplementary material.

\subsection{Computational Efficiency}

\begin{table}[t]
\centering
\caption{Computational efficiency comparison of our approach against state-of-the-art geo-localization methods. All GPU measurements conducted on NVIDIA RTX 3080 with batch size 32 to match the scenario in TransGeo \cite{zhu2022transgeo}.}
\label{tab:prior_art_comparison}
\resizebox{0.6\columnwidth}{!}{%
\begin{tabular}{lccc}
\toprule
Model & \begin{tabular}[c]{@{}c@{}}GPU\\Memory (GB)\end{tabular} & \begin{tabular}[c]{@{}c@{}}Parameter\\Count (M)\end{tabular} & \begin{tabular}[c]{@{}c@{}}GPU Inference\\Time (ms)\end{tabular} \\
\midrule
VAE-Transformer & \textbf{2.12} & \textbf{19.2} & \textbf{37} \\
VIGOR (SAFA \cite{shi2019spatial}) & 10.82 & 29.4 & 111 \\
TransGeo & 9.85 & 44.8 & 99 \\
\bottomrule
\end{tabular}%
}
\end{table}

Table \ref{tab:prior_art_comparison} presents the computational efficiency comparison of our VAE-Transformer model against VIGOR and TransGeo across key efficiency metrics. Our VAE-Transformer significantly outperforms state-of-the-art alternatives in all efficiency metrics. Our approach delivers superior computational efficiency while achieving better localization accuracy. The trained model is just 77MB in size, making it highly portable and deployable on memory-constrained devices. Most impressively, our model processes inputs 3× faster during inference compared to both competitors.

The practical advantage becomes even more evident in real-world deployment scenarios. Our model achieves real-time performance (10.9 FPS) even on the Jackal robot's modest onboard CPU (Intel i5-4570T) without requiring specialized hardware acceleration. This efficiency enables direct deployment on resource-constrained robotic platforms and embedded systems in autonomous vehicles.

\section{Discussion}

\subsection{Cognitive Model Outperforming Engineering Solutions}

This study builds upon a simple yet profound observation about biological cognitive abilities: humans and animals can accurately memorize and localize within their environments without relying on explicit coordinate systems or external reference signals. By implementing these biologically-inspired principles, we have created a localization system that not only achieves state-of-the-art performance, but also demonstrates remarkable computational efficiency, enabling real-time operation on resource-constrained platforms.

Our results confirm that the VAE-Transformer can effectively model the spatial reality of complex environments, from a university campus with pedestrian traffic to an urban downtown with both pedestrian and vehicular movement. Significantly, this approach outperforms engineering solutions to the same problem, which typically rely on computational techniques of search and retrieval rather than generative understanding. In some scenarios, our model even matches the performance of physical measurement devices (smartphone GPS), despite the latter having access to global positioning satellites and specialized hardware. This finding demonstrates that given sufficient training data and appropriate parameter capacity relative to environment size, a generative sequential model can succeed in modeling the complex reality of a fixed location. 

Our reconstruction ablation analysis further reinforce this conclusion, showing that the reconstruction objective—which forces the model to generate aligned satellite views from street-level observations—is critical for accurate localization in complex environments. This mirrors biological spatial cognition, where transforming between perspectives helps maintain consistent spatial representations across varying sensory inputs.

The success of our cognitive modeling approach suggests that for certain spatial intelligence tasks, mimicking the brain's information processing strategies may be more effective than engineering solutions designed from first principles. This perspective invites a broader reconsideration of how we approach complex computational problems across various domains, potentially leading to more efficient and robust AI systems that draw inspiration from nature's time-tested designs.

\subsection{Generalization vs. Specialization}

Our research challenges the conventional wisdom that favors maximum generalization across spatial domains. We propose inverting the traditional approach: rather than spatial generalization, we advocate for spatial specialization with temporal generalization.

This approach acknowledges the i.i.d. assumption fundamental to machine learning. Attempts to use models trained in one location to perform in another essentially measure correlation between feature distributions. Rather than fighting this constraint, our approach embraces it by clearly defining operational boundaries for each model.

The practical advantage is clear in our results: a lightweight (77MB) model specialized for a specific area achieves GPS-rivaling precision while occupying just 5$\times$ more storage than the raw satellite imagery of the same area (15MB). This efficiency enables practical deployment scenarios such as map service providers offering a ``database of STRMs" instead of merely satellite imagery.

While our approach requires more data from each specific location, this challenge can be addressed through crowdsourcing video data from millions of autonomous vehicles that are already collecting and transmitting footage for centralized processing, creating a rich dataset for training location-specialized models with minimal additional infrastructure investment.

This doesn't render generalized models obsolete. Models could still generalize across similar environmental categories (e.g., North American urban environments). However, we should be realistic about generalization limits, particularly when environments differ substantially.

A complete system could be hierarchical: a map of specialized models covering defined regions, potentially supplemented by broader models for similar environment types. This approach better aligns with both theoretical constraints of machine learning and practical requirements of real-world deployment.

\subsection{STRM Applications: Location-Specific Autonomous Driving}

A compelling application of STRMs is in location-specific autonomous driving. Current autonomous systems rely on general policies without location-specific context, causing them to struggle with unique environmental features like subtle road dips that human drivers navigate effectively through repeated exposure. This work demonstrates that STRMs can learn the association between location-specific visual cues and coordinates, suggesting they could similarly learn to predict appropriate location-specific driving actions, by training on crowdsourced human driving data.

A scalable implementation could be STRMs of different sizes being deployed based on location complexity, forming a ``map of specialized spatial intelligence." Autonomous Driving Companies with their extensive data collection capabilities, could implement location-specific STRMs as specialized modules working alongside their main autonomous driving systems. These modules would provide tailored behaviors for challenging locations, potentially addressing edge cases that currently confound autonomous systems.

\subsection{Conclusion and Future Work}

Our research demonstrates that a cognitive model inspired by biological spatial intelligence can outperform traditional engineering solutions for localization. Future work should investigate the performance of our approach across day and night time and extended time periods to evaluate how environmental changes such as seasonal variations, construction activities, and shifting lighting conditions affect localization accuracy. Developing mechanisms to maintain performance despite these temporal variations would significantly enhance the practical applicability of STRMs.

\section*{Acknowledgments}
This work was supported by the Air Force Office of Scientific Research (AFOSR) Contract No. FA9550-19-1- 0306, and by the National Institute of Neurological Disorders and Stroke (NINDS) Award No. R01NS135850.

\bibliography{main}  
\bibliographystyle{ieeenat_fullname}

\end{document}